\documentclass[lettersize,journal]{IEEEtran}
\usepackage{amsmath,amsfonts}
\usepackage{algorithmic}
\usepackage{algorithm}
\usepackage{array}
\usepackage[caption=false,font=normalsize,labelfont=sf,textfont=sf]{subfig}
\usepackage{textcomp}
\usepackage{stfloats}
\usepackage{url}
\usepackage{verbatim}
\usepackage{graphicx} 
\usepackage{subfig}
\usepackage{cite}
\usepackage{multirow}
\usepackage{xspace}
\newcommand{\name}[0]{CPES\xspace}
\usepackage{xcolor}

\begin{document}

%\title{Selecting Class-Relevant Patch Embeddings \\ for Few-Shot Learning} 
\title{Class-relevant Patch Embedding Selection for Few-Shot Image Classification}

\author{Weihao Jiang, Haoyang Cui, Kun He$^*$,~\IEEEmembership{Senior Member,~IEEE}
        % <-this % stops a space

\thanks{Weihao Jiang, Haoyang Cui and Kun He are with the School of Computer Science, Huazhong University of Science and Technology, Wuhan, China 430074. Corresponding author: Kun He, E-mail: brooklet60@hust.edu.cn.}
%\thanks{This paper was produced by the IEEE Publication Technology Group. They are in Piscataway, NJ.}% <-this % stops a space
\thanks{This work is supported by National Natural Science Foundation of China (U22B2017).}
\thanks{}}

\markboth{}%
{Shell \MakeLowercase{\textit{et al.}}: A Sample Article Using IEEEtran.cls for IEEE Journals}

\IEEEpubid{}

\maketitle

\begin{abstract}
Effective image classification hinges on discerning relevant features from both foreground and background elements, with the foreground typically holding the critical information. 
While humans adeptly classify images with limited exposure, artificial neural networks often struggle with feature selection from rare samples. 
To address this challenge, we propose a novel method for selecting class-relevant patch embeddings. Our approach involves splitting support and query images into patches, encoding them using a pre-trained Vision Transformer (ViT) to obtain class embeddings and patch embeddings, respectively. Subsequently, we filter patch embeddings using class embeddings to retain only the class-relevant ones. For each image, we calculate the similarity between class embedding and each patch embedding, sort the similarity sequence in descending order, and only retain top-ranked patch embeddings.
%, and discard the remaining ones with low similarity values. 
By prioritizing similarity between the class embedding and patch embeddings, we select top-ranked patch embeddings to be fused with class embedding to form a comprehensive image representation, enhancing pattern recognition across instances. Our strategy effectively mitigates the impact of class-irrelevant patch embeddings, yielding improved performance in pre-trained models.  
Extensive experiments on popular few-shot classification benchmarks demonstrate the simplicity, efficacy, and computational efficiency of our approach, outperforming state-of-the-art baselines under both 5-shot and 1-shot scenarios. 
\end{abstract}


% ---------------------------------------------------------------------------------
\section{Introduction}
% \IEEEPARstart{T}{his} file is intended to serve as a ``sample article file''
Deep learning has achieved remarkable advancements in computer vision and multimedia systems.
However, its dependency on extensive annotated data poses challenges when confronted with novel categories with very few labeled examples. Few-shot learning methods~\cite{siamese,one-shot,adversarial,Attribute-Guided,PrototypicalNetwork,SgVA-CLIP,Graph}  have emerged as a solution to these challenges by leveraging knowledge acquired from base categories to generalize to novel ones, thereby mitigating issues stemming from data scarcity. 
Unlike traditional methods requiring extensive datasets, few-shot learning operates on a small number of labeled samples, rendering it more practical for real-world applications by 
%avoiding time-consuming and expensive data annotation processes. 
circumventing laborious and costly data annotation processes. 
Drawing inspiration from human's rapid learning through effective utilization of prior knowledge, few-shot learning aims to resolve this problem through knowledge transfer, with metric-based meta-learning approaches~\cite{RelationNetwork,PrototypicalNetwork,Meta-Baseline}, showcasing simple yet effective characteristics. This paradigm significantly reduces overhead associated with applying deep learning to multimedia systems and has garnered increasing attention in the literature. %research community.

\begin{figure}[t!]
  \subfloat[%A pair of 
  Images of distinct %different 
  categories]{\includegraphics[width=0.48\textwidth]{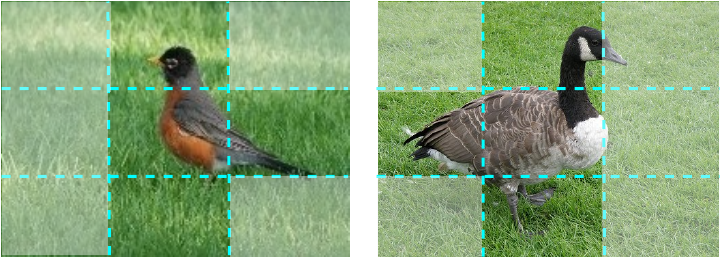}}
   \newline
  \subfloat[%A pair of 
  Images of the same category]{\includegraphics[width=0.48\textwidth]{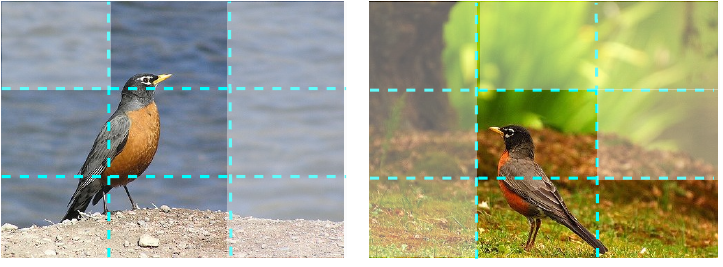}}
\caption{Illustration on the patch regions. The highlighted regions contain key semantics consistent with global information that corresponds to semantics of image labels, while low-transparency regions contain semantics not relevant to the global information. %In this paper, global information refers to the semantics of image labels, such as ``bird" and %``duck"
%``goose" in the figures.
}
\label{fig:local}
\end{figure}

\IEEEpubidadjcol

Few-shot learning methods based on metric learning typically involve three fundamental steps: 
%Firstly, feature extraction from all query and support images. Secondly, calculation of distances between the query image and the center of each support image class using a designated metric. Finally, assignment of labels to the query image through a nearest neighbor search. 
feature extraction from query and support images, computation of distances between the query image and the center of each support image class using a designated metric, and label assignment to the query image via nearest neighbor search.
However, in the context of few-shot image classification, the reliance on only a few images to represent a new category poses a significant challenge. On one hand, the model encounters difficulties in identifying the defining characteristics that determine the category of an image due to the limited availability of labeled images for the new category and the extremely limited number of associated entities. On the other hand, potential ambiguity arises when an entity present in the training image, not covered by the training class, aligns with the entity expected to be represented by the new class during the testing phase.

To tackle these challenges,  researchers have introduced various approaches emphasizing class-relevant regions or augmenting the relevance of region features. 
SAML~\cite{saml} employs a ``collect-and-select" strategy, employing attention techniques to select semantically relevant pairs and assign them higher weights. CAN~\cite{CAN} generates cross attention maps between class features and query sample features to highlight target regions. 
CTX~\cite{CTX} proposes a Transformer-based neural network architecture to establish coarse spatial correspondence between queries and labeled images. 
DeepEMD~\cite{deepemd} achieves semantic alignment by minimizing the earth movers' distance between image pairs. 
FewTURE~\cite{FewTrue} selects the most informative patch embeddings through online optimization during inference. 
Recently, CPEA~\cite{cpea} integrates patch embeddings with class-relevant embeddings to enhance their class relevance.

While these methods show promise in mitigating interference, they also face notable limitations. Firstly, the scarcity of labeled images for new classes and the limited number of class-relevant entities present challenges in accurately positioning and aligning objects in natural images, particularly in the presence of intra-class variations and excessive background clutter. Secondly, when relying on regional feature measurements, these methods consider all local regions without effectively filtering out class-irrelevant regions, leading to inaccurate measurements.

As depicted in Figure~\ref{fig:local}, the %target region
focal region (foreground) within an image typically occupies only a fraction, while the remainder comprising  interfering elements such as the background. When comparing two images from distinct categories, although noticeable distinctions  %in the target region
exist in the focal region, %it constitutes a small proportion
its contribution constitutes  a minor fraction of the overall image. Consequently, when evaluating measurements across all patches, the discriminative capacity  
%of the target region is lower compared to that of the background region. 
of the focal region  is diminished compared to that of the background. 
This discrepancy can lead to the misclassification of both images into the same category. Similarly, even two images from identical categories may be erroneously identified as belonging to different categories due to the interference of background factors.

%In this paper, we address the aforementioned challenges with a novel approach, presenting the Selecting Class-Relevant Patch Embeddings (\name) method.
In this paper, we address the aforementioned challenges with a novel approach, the Class-relevant Patch Embedding Selection (\name) method. \name effectively eliminates interference from class-irrelevant regions while preventing supervision collapse.
Specifically, our method leverages a Vision Transformer (ViT) architecture as the feature extractor,%. Instead of supervised pretraining, we employ self-supervision pretraining with Masked Image Modeling (MIM) as a pretext task, generating semantically meaningful patch embeddings.
 utilizing self-supervised pretraining with Masked Image Modeling (MIM) as a pretext task so as to generate semantically meaningful patch embeddings. 
We then partition the image into disjoint patches and feed them into the pre-trained feature extractor to obtain class embedding and patch embeddings, that represent global information and local information, respectively. 
%Some patch embeddings may be irrelevant to the class of interest, causing interference in measuring the similarity of image pairs. To address this, we introduce a method to select class-relevant patch embeddings. Finally, we compute a dense score matrix between class-relevant patch embeddings across images to quantify the similarity between paired images.
%To mitigate the interference of irrelevant patch embeddings, we introduce a method to select class-relevant ones. 
We then pick top relevant patches to mitigate the interference of irrelevant patch embeddings for each image. 
In the end, for each pair of images, we compute a dense score matrix between their class-relevant patch embeddings to quantify similarity.
%across images to quantify similarity between paired images.
% Following this, we employ a linear strategy to combine the class embedding with the selected patch embeddings, resulting in fused patch embeddings. Subsequently, for each pair of images, we compute a dense score matrix using the fused class-relevant patch embeddings. This score matrix is then fed into a multi-layer perceptron to calculate the similarity score. 

Our contributions can be summarized as follows:

$\bullet$ We address the background interference in few-shot scenarios from a novel perspective, demonstrating effective mitigation without requiring localization and alignment mechanisms.

$\bullet$ We introduce \name, a novel approach that selects class-relevant patch embeddings and quantifies similarity between the embeddings across images in a dense manner, thereby enhancing model generalization.

$\bullet$ Visualizations demonstrate the efficacy of \name in selecting class-relevant patch embeddings, %Extensive experiments conducted on four prominent benchmark datasets showcase that \name outperforms state-of-the-art methods.
and extensive comparisons showcase the superiority of our method. 

\section{Related Works}
% few-shot methods： metric , optimization
% SSL in fsl
% Attention
In this section, we start by categorizing various approaches to few-shot learning and proceed to conduct a comparative analysis between our method and existing ones. Subsequently, we outline several ViT pre-training methodologies, ultimately selecting Masked Image Modeling as our preferred pre-training technique. Lastly, we delve into the utilization of attention mechanisms, highlighting the distinctive  aspects of our selection methodology.

\subsection{Few-shot Learning}
In recent years, there has been a significant body of research dedicated to  few-shot learning. Broadly, these efforts can be categorized into two main branches: meta-learning and transfer learning.
Certainly, most of the existing few-shot learning methods rely on the meta-learning framework, which can be further divided into two types: optimization-based and metric-based. 

\textbf{Optimization-based methods.} Optimization-based methods aim to learn a good model initialization %learn the model parameters
such that the model can be optimized quickly when encountering new tasks. 
As a representative work of this category, MAML~\cite{MAML} obtains good initialization parameters through internal and external training in the meta-training stage. 
MAML has inspired many follow-up efforts, such as ANIL~\cite{ANIL}, BOIL~\cite{BOIL}, and LEO~\cite{LEO}.

\textbf{Metric-based methods.} Metric-based FSL methods embed support images and query images into the same space and classify query images by calculating the distance or similarity. 
Using different metric methods, various models have been derived, such as Matching Networks~\cite{MatchingNetwork}, Prototypical Networks~\cite{PrototypicalNetwork}, MSML~\cite{msml} and DeepEMD~\cite{deepemd}. Simultaneously, to garner more distinctive features, researchers have started treating the entire task holistically and devising external modules to enhance the features.
CTM~\cite{CTM} introduces a Category Traversal Module that traverses the entire support set at once, discerning task-relevant features based on both intra-class commonality and inter-class uniqueness in the feature space. CAN~\cite{CAN} generates cross-attention maps for each pair of support feature map and query sample feature map to emphasize target object regions, enhancing the discriminative nature of the extracted features. CTX~\cite{CTX} proposes to learn spatial and semantic alignment between CNN-extracted query and support features using a Transformer-style attention mechanism. ATL-Net~\cite{ATL-Net} introduces episodic attention calculated by a local relation map between the query image and the support set, adaptively selecting important local patches within the entire task. RENet~\cite{RENet} combines self-correlational representation within each image and cross-correlational attention modules between images to learn relational embeddings.

Recently, researchers have started incorporating ViT into few-shot learning scenarios. 
The PMF~\cite{P>M>F} method consists of three phases. Firstly, it utilizes a pre-trained Transformer model with external unsupervised data. Then, it simulates few-shot tasks for meta-training using base categories. Finally, the model is fine-tuned using scarcely labeled data from test tasks.
HCTransformer~\cite{HCTransformer} employs hierarchically cascaded Transformers as a robust meta feature extractor for few-shot learning.
It utilizes the DINO~\cite{DINO} learning framework and maximizes the use of annotated data for training. Additionally, it enhances data efficiency through attribute surrogates learning and spectral tokens pooling.
FewTURE~\cite{FewTrue} adopts a completely transformer-based architecture and learns a token importance weight through online optimization during inference.
CPEA~\cite{cpea} also adopts a pre-trained ViT model and integrates patch embeddings with class-aware embeddings to make them class-relevant.

\textbf{Transfer learning based methods.} Few-shot learning methods based on the transfer learning framework yield comparable results compared with meta-learning approaches. 
They first employ a simple scheme to train a classification model on the overall training set, then remove the classification head and retain the feature extraction part, and train a new classifier based on the support set from testing data.  
Representative works include Dynamic Classifier~\cite{dynamic}, Baseline++~\cite{baseline++} and RFS~\cite{RFS}.

Our proposed method falls into the category of metric-based meta-learning methods. 
In contrast to existing metric-based learning approaches that directly utilize all features, our method exploits the semantic correlation between global and local features to emphasize the target object. 
Although some methods~\cite{CTM,CAN,CTX,ATL-Net,RENet,FewTrue,cpea} also consider to highlight helpful local regions, they rely on all local features and weight these features through complex external module networks. 
In contrast, our approach solely
employs top class-relevant patch embeddings based on the similarity sequence between class embedding and patch embeddings, and gains superior performance. 
%utilizing self-supervised pre-trained ViT model. 
% achieves superior performance with solely selected class-relevant patch embeddings utilizing self-supervised pre-trained ViT model. 
% As demonstrated in the followup experiments, our method outperforms these approaches by a significant margin.

\subsection{Self-supervised Pretraining ViT}
Vision Transformers have shown superior performance compared to traditional architectures~\cite{dsp,HRTransNet}. However, their effectiveness heavily relies on a vast amount of labeled data, which may not always be feasible or sustainable for supervised training. Thus, implementing a self-supervised approach for Vision Transformers can make them powerful and more applicable to downstream problems.
BEiT~\cite{BEiT} introduces the concept of Masked Image Modeling as a self-supervised training task for training ViT.
%During pre-training, BEiT tokenizes the original image, randomly masks image blocks, and feeds the masked image into the encoder. The main goal of pre-training is to restore the masked image block based on the unmasked counterpart.
SimMIM~\cite{SimMIM} suggests that randomly masking image parts and increasing the resolution of image blocks can yield satisfactory results.
%It proposes that the optimal performance is achieved when the image block resolution is 32$\times$32.
In contrast to BEiT, MAE~\cite{MAE} simplifies the training logic by using random masks to process input image blocks and directly reconstructing the masked image blocks for training. 
%MAE incorporates two key designs. First, it employs an encoder-decoder architecture with an asymmetric structure, where the encoder only considers non-masked image blocks, along with a lightweight decoder design. Second, it covers a majority of the image blocks, such as using a mask probability of 75\%, to obtain a more meaningful self-supervised training task.
DINO~\cite{DINO} presents a straightforward self-supervised approach for ViT, which can be viewed as a form of knowledge distillation without the use of labels. 
The DINO framework involves predicting the output of a teacher network, constructed with a momentum encoder, by employing a standard cross-entropy loss.

In our approach, we leverage a Vision Transformer architecture as an encoder, chosen for its patch-based nature. Simultaneously, to address the issue of supervision collapse, we implement self-supervised training, employing Masked Image Modeling as a pretext task. Moreover, a considerable number of Transformer models pre-trained using self-supervised methods exist. To expediently assess the extensibility of our model, we incorporate certain pre-trained model parameters when utilizing the ViT-Base model, specifically those obtained through pre-training with the MAE method.

\subsection{Attention Mechanism}
Attention mechanisms have been successful in highlighting crucial local regions to extract more discriminative features, notably in various computer vision tasks such as image classification, image captioning, and visual question answering. In few-shot image classification, prior works~\cite{ATL-Net,CTM,FewTrue} propose weight generator modules based on attention mechanisms to enhance model representation. These modules derive critical weights for each local feature to emphasize the focal %target
region. However, these methods rely on weighted aggregations of all local features, which, although effective in highlighting focal %target 
region, may still be susceptible to background interference and other factors. Moreover, they localize critical regions in test images solely based on training class priors, limiting generalization to unseen classes. In contrast to existing approaches, our paper introduces a novel method for selecting class-relevant patch embeddings. This approach enables the identification of crucial regions in the target image and enhances feature discriminability.

\section{Methodology}
In this section, we begin by providing the definition of few-shot image classification task. Consequently, we outline the framework of our proposed method. Then, we elaborate on the module concerning our class-relevant patch embedding selection in detail, and explain the classification method based on the chosen patch embeddings.

\subsection{Problem Definition}
Few-shot image classification is primarily concerned with the $N$-way $K$-shot problem, where $N$ denotes the number of categories and $K$ represents the number of instances within each category. Typically, $K$ is relatively small, often set to values like 1 or 5.

Datasets designed for few-shot learning typically comprise three distinct parts: training set, validation set, and test set. Notably, they feature non-overlapping categories, meaning that images in the test set are entirely unseen during training and validation. This lack of overlap poses a significant challenge for few-shot learning. Typically, all three datasets contain numerous categories and examples. To emulate the conditions of few-shot learning, researchers adopt an episode training mechanism~\cite{MatchingNetwork}. In this mechanism, episodes are randomly sampled from the datasets, each consisting of a support set and a query set. Both the support set and query set have identical labels, but their samples do not overlap. The support set contains a sparse selection of labeled samples, acting as the few-shot training data, while the query set is used for evaluation.

During the training phase, a large number of episodes sampled from the training set are utilized to update the model parameters until reaching convergence. For validation and testing, episodes from the validation and test sets are employed to ensure that the test conditions mirror real-world scenarios.

\begin{figure*}[!t]
\begin{center}
%\fbox{\rule{0pt}{1in} \rule{.9\linewidth}{0pt}}
 % \includegraphics[width=0.9\linewidth]{svrg-3-2.svg}
  \includegraphics[width=1.0\linewidth]{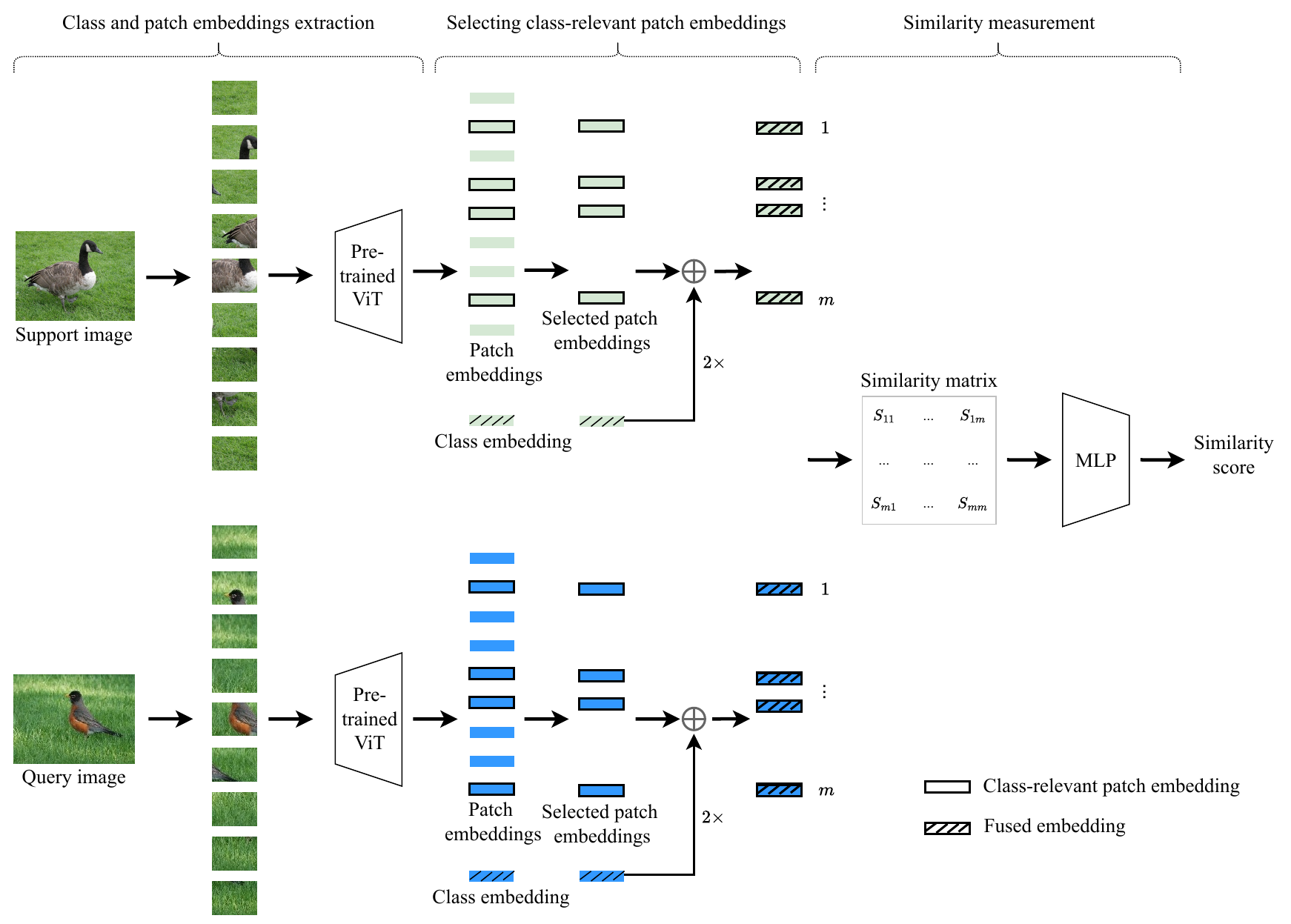}
\end{center}
   \caption{
   %Illustration of the \name processing pipeline for few-shot learning task. 
   The processing pipeline of \name. 
   %Here we use 2-way 1-shot as an example. 
   %The support and query images are firstly patched and then the pre-trained ViT is used to encode the input images. After getting the class embedding and patch embeddings, we use the similarities between class embedding and patch embeddings to select top class-relevant patches. Then, we fuse class embedding with the selected patch embeddings to obtain new patch embeddings. Based on these new patch embeddings, a similarity metric is obtained between each pair of image. Finally, the similarity metric is flattened and fed into a multi-layer perceptron to obtain a similarity score.
   Support and query images are patched and then encoded with a pre-trained ViT. Patch embeddings are compared with class embedding to select top relevant patches, which are then fused with class embeddings to create new embeddings. A similarity matrix is calculated based on these embeddings, which in the end is flattened and fed into a multi-layer perceptron to generate the similarity score.
   }
\label{fig:framework}
\end{figure*}

\subsection{The Proposed Framework}
The framework of our \name model, as depicted in Figure~\ref{fig:framework}, leverages the Vision Transformer as the feature extractor. 
Each input image is first partitioned into non-overlapping patches, and  encoded using the pre-trained ViT~\cite{ViT}, yielding both class embedding and patch embeddings. 
While the class embedding captures global image information, patch embeddings capture local details specific to each patch. 
% . These patches, from both the support set and the query set, are then encoded using the pre-trained ViT~\cite{ViT}, resulting in class embeddings and patch embeddings. The class embedding capture global information of the entire image, whereas the patch embeddings represent local information specific to each patch.
%Then, both the class embedding and patch embeddings are fed to our patch embedding selection module, in which we utilize the class embedding to select class-relevant patch embeddings, such that the class-irrelevant patch embeddings are eliminated for image representations. 
Subsequently, in our patch embedding selection module, the class embedding is employed to filter out class-irrelevant patch embeddings, enhancing image representations by focusing on class-relevant regions.
%By screening class-relevant patch embeddings, the impact of class-irrelevant regions in the background on image representation can be reduced, so that the retained features can be more focused on the focal region. 
This process mitigates the influence of irrelevant background regions, sharpening the emphasis on the key image features. 
We then fuse the class embedding with each selected patch embedding to create new patch embeddings%. The new patch embeddings containing local features will also bear the imprint of global information. 
, enriching local features with global context. For instance in an image of a cat, where `cat' represents the class embedding and `ear' denotes one patch embedding, the fused `ear' patch embedding becomes more distinct and representative. % and distinguishable. 
For every pair of images drawn from the support and query sets, 
we calculate a similarity matrix based on their fused patch embeddings. 
Finally, a multi-layer perceptron (MLP) layer is employed to calculate the similarity score from the flattened  similarity matrix.
%we calculate the similarity value between each pair of patches of the query image and the support set image to obtain a similarity matrix. 
%Finally, we flatten the similarity matrix and feed it into a multi-layer perceptron (MLP) network layer to aggregate all the similarities and obtain the final similarity score between the query and support images.

% Unlike the traditional methods, our method is not based on all patch embeddings for measurement.  However, it uses the feature of the class embedding, which represents global information, to select the class-relevant features in the local features to eliminate the impact of class-irrelevant patch features on image representation.  Although some previous methods learn the weights of different patches to distinguish which patches are more valuable, this still cannot eliminate the negative impact of class-irrelevant patches.  At the same time, this method usually requires a network module to learn the weight parameters, which increases the number of parameters and makes generalization challenging to guarantee.  Our method does not need additional network modules to learn the weight parameters.  However, it only needs to be sifted according to the similarity between class and patch embeddings, and only one needs to select the number of patches as a hyperparameter.  The measurement based on the selected patch embeddings can make the network more focused on the target region.
Our approach differs from traditional methods in that it does not rely on all patch embeddings for measurement. Instead, it leverages the global information of class embedding to identify class-relevant features within the local patch features, thereby mitigating the impact of class-irrelevant patch features on image representation. 
Although previous methods have attempted to learn the weights of different patches to discern their value, they have failed to completely eliminate the negative impacts of class-irrelevant patches. Moreover, these methods often require an additional network module to learn the weight parameters, resulting in an increased number of parameters and posing challenges for generalization. 
In contrast, our method does not necessitate extra network modules for weight parameter learning. Rather, it only requires sifting based on the similarity between class and patch embeddings, with the selection of the number of patches as the sole hyperparameter. %The measurement based on the selected patch embeddings enables the network to focus more precisely on the focal region.
The utilization of selected patch embeddings allows the network to precisely focus on the focal region.

\subsection{Class-relevant Patch Embedding Selection}
In a typical single few-shot task, the support set usually consists of five classes, where $N$ is commonly set to five. Each support category contains $K$ shots, and the average of the embedding values is taken as the prototype embeddings. Following the encoding process, we obtain five sets of combinations, namely class embedding and patch embedding pairs:
\begin{align}\label{e1}
    P_i &= [class_{i}^p,patch_{i,1}^p,patch_{i,2}^p,...,patch_{i,M}^p],\\
    &i\in [1,5], \nonumber
\end{align}
% $i\in [1,5]$,
where $M$ is the number of patches.

For a query image, we have a combination of class embedding and patch embeddings:
\begin{align}\label{e2}
    Q  = [class^q,patch_{1}^q,patch_{2}^q,...,patch_{M}^q],
\end{align}
where $M$ is the number of patches.
% Where, $class$ is the class embedding representing the global information of the image, and $patch$ is the patch embedding representing the feature of local region in the image. Obviously, the semantics of these patch embeddings may be independent of the class to which the image belongs, and the spatial location of the patch embeddings associated with the class of interest is unknown in advance. Due to the large intra-class variation and background clutter, the measurement of using all patch embeddings directly for image classification is far from satisfactory. Therefore, it is more reasonable to first select the patches that are most relevant to the category and then carry out subsequent operations.
Here $class$ is the class embedding representing the global information of the image, and $patch$ is the patch embedding representing the feature of the local region in the image. The semantics of these patch embeddings may be independent of the class to which the image belongs, and the spatial location of the patch embeddings associated with the class of interest is unknown in advance. Due to the large intra-class variation and background clutter, the measurement of using all patch embeddings directly for image classification is far from satisfactory. Therefore, it is more reasonable to filter out the most relevant patches to the category and then carry out subsequent operations.

For an image, we obtain its class embedding and patch embeddings respectively, namely:
\begin{align}\label{e3}
    P = [class,patch_{1},patch_{2},...,patch_{M}], 
\end{align}
and then we calculate the similarity between class embedding and each patch embedding using the following formula:
\begin{align}\label{e4}
    &r_i = cos(class,patch_{i}),\\
    &i\in [1,2,...,M], \nonumber
\end{align}
where $cos(\cdot)$ indicates the cosine similarity. Other similarity or distance functions can certainly be employed.
The similarity sequence of an image between its class embedding and patch embeddings is:
\begin{align}\label{e5}
    R = [r_1,r_2,r_3,...,r_M].  
\end{align}

% For the obtained similarity sequence, we sorted it in descending order, selected the first $m$ patch embeddings, obtained the index value of patch in all patches through equation~\eqref{e6} , and retained these patch embeddings with high similarity to class according to the index.
The obtained similarity sequence is then sorted in descending order. We select the top $m$ patch embeddings based on the sorted sequence. To obtain the index value of a patch among all patches, equation~\eqref{e6} is utilized. We retain those patch embeddings that exhibit high similarity to the class, as determined by their corresponding index values.
\begin{align}\label{e6}
    &id_i = topK(R,m),\\
    &i\in [1,2,...,m], 1\le id_i \le M. \nonumber
\end{align}

\begin{figure*}[t!]
  \subfloat[Task 1 without \name]{\includegraphics[width=0.25\textwidth]{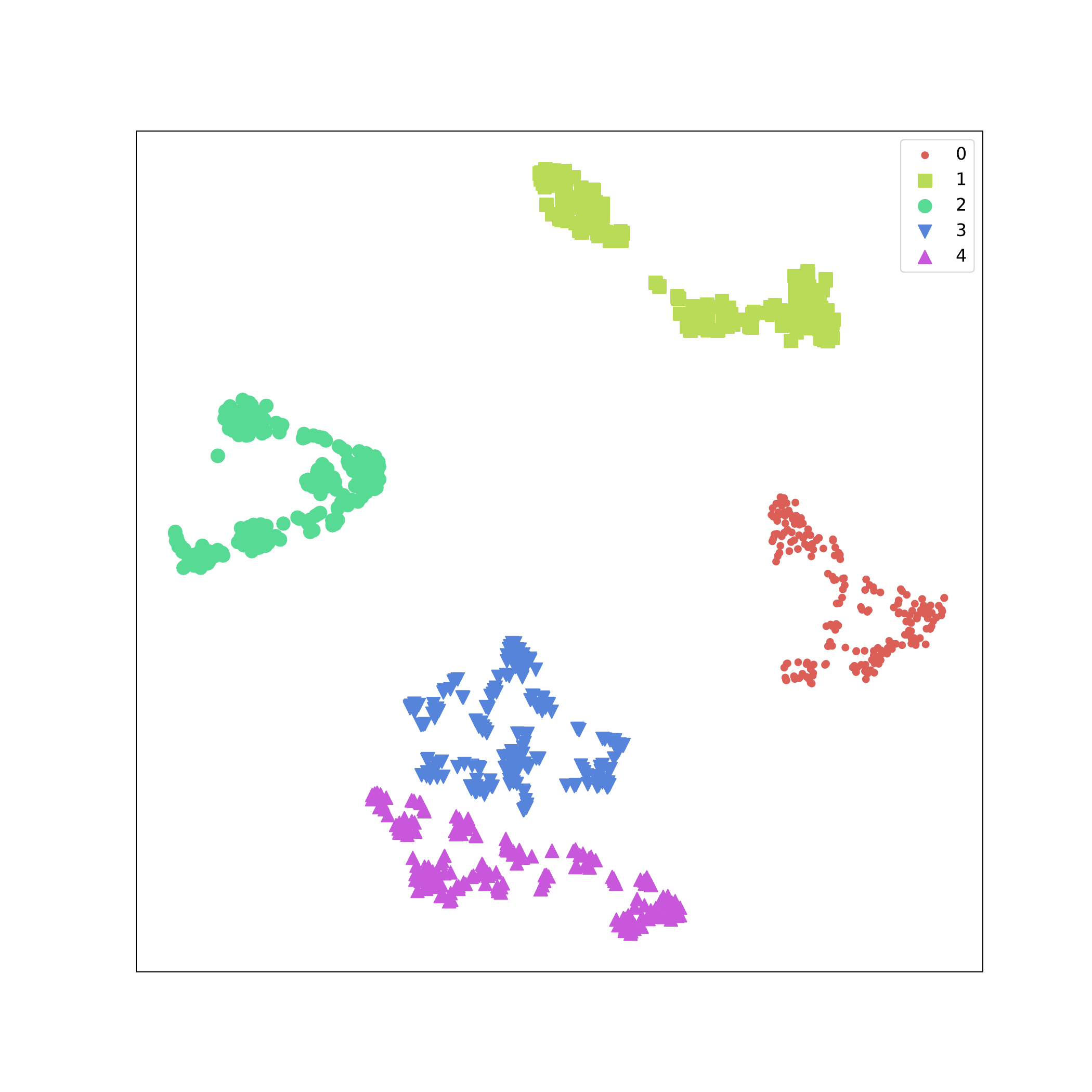}}
 \hfill 	
  \subfloat[Task 2 without \name]{\includegraphics[width=0.25\textwidth]{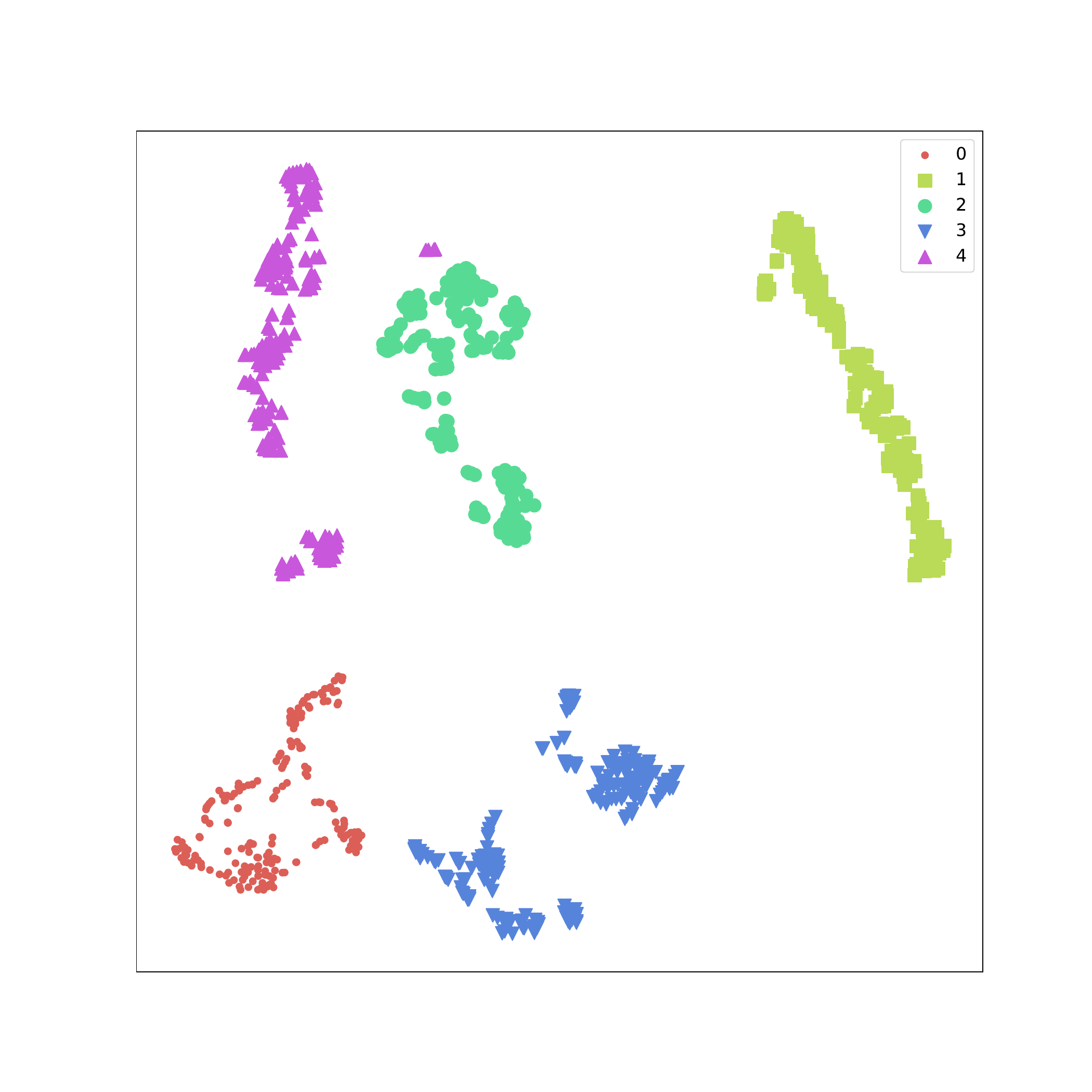}}
 \hfill	
  \subfloat[Task 3 without \name]{\includegraphics[width=0.25\textwidth]{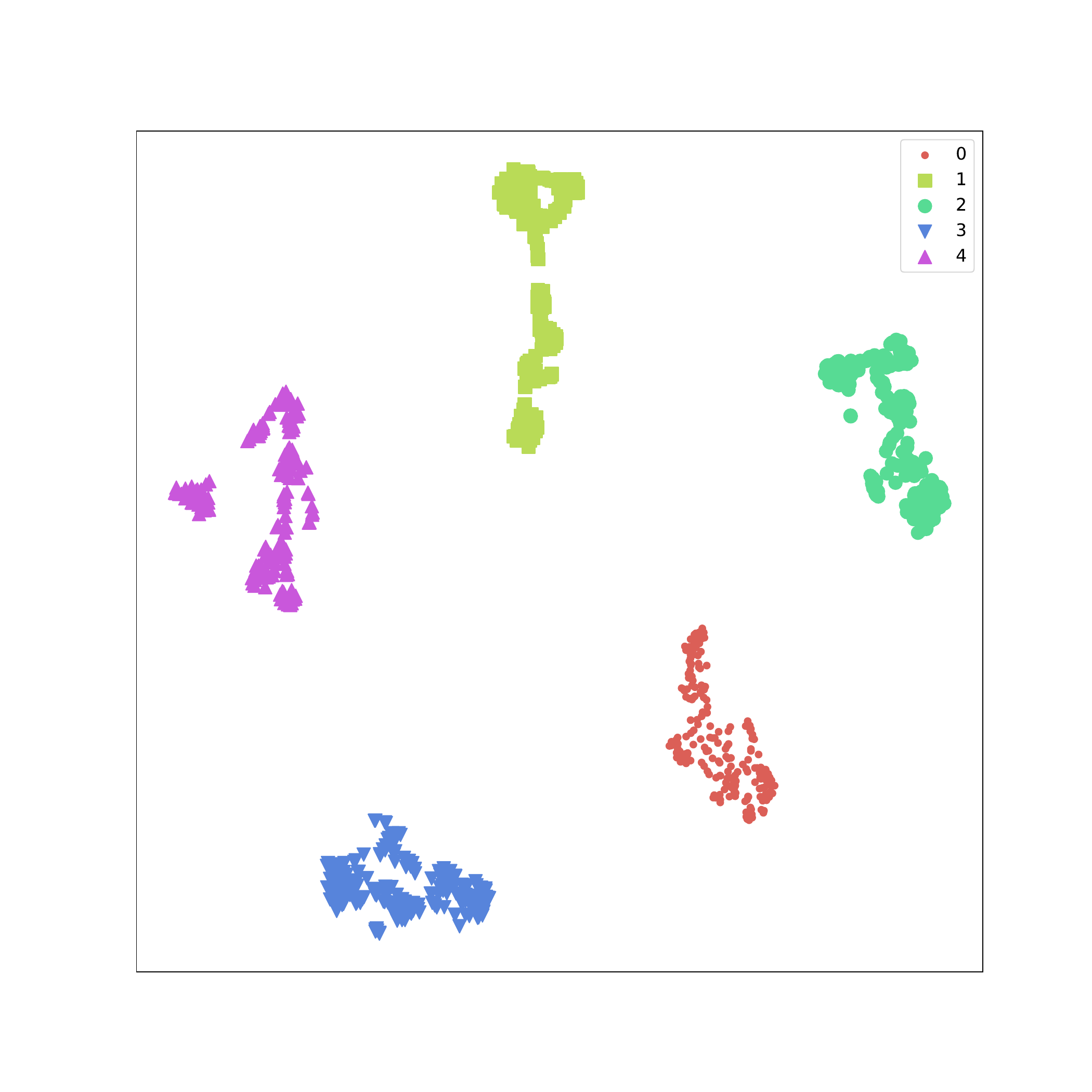}}
  \hfill	
  \subfloat[Task 4 without \name]{\includegraphics[width=0.25\textwidth]{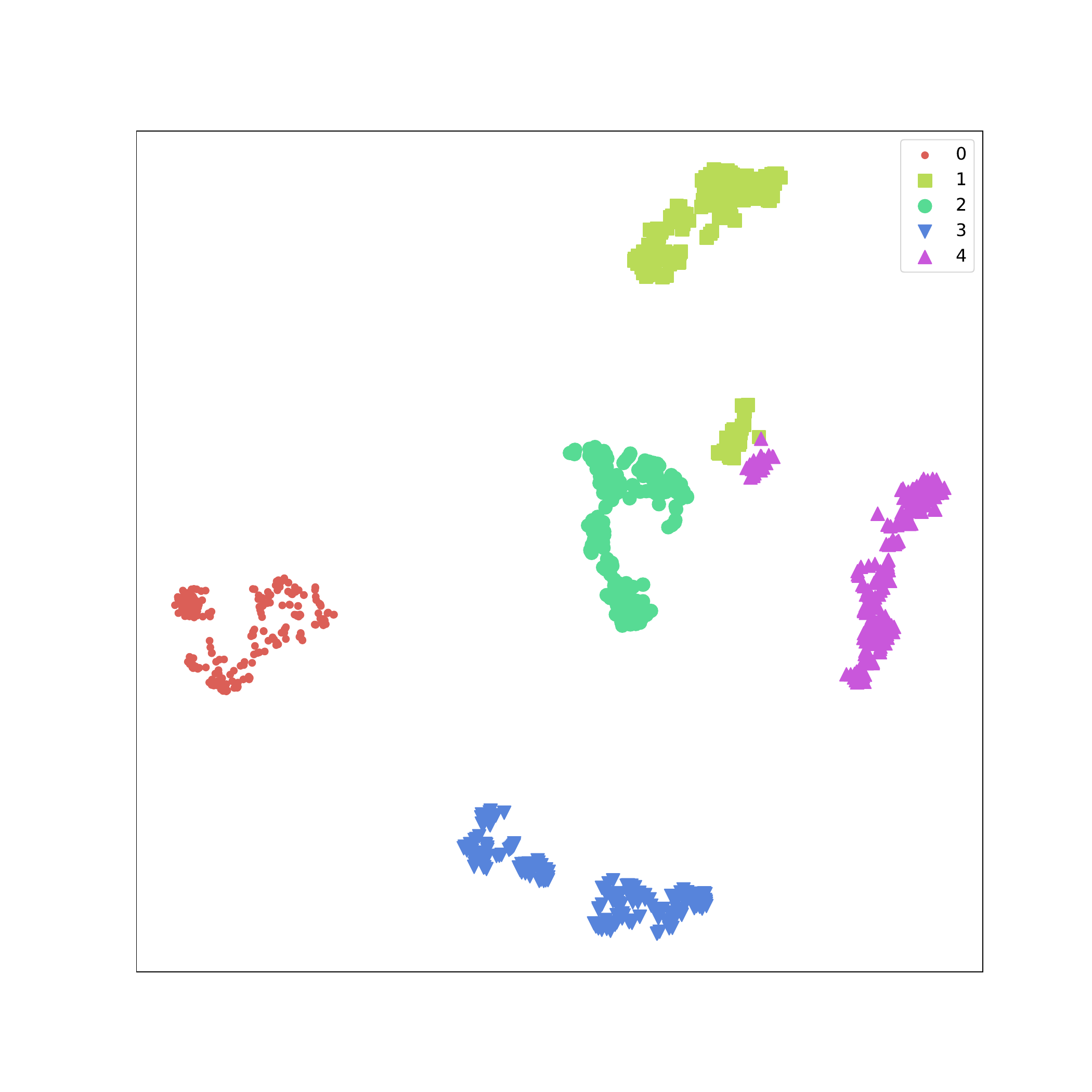}}
  \newline
  \subfloat[Task 1 with \name]{\includegraphics[width=0.25\textwidth]{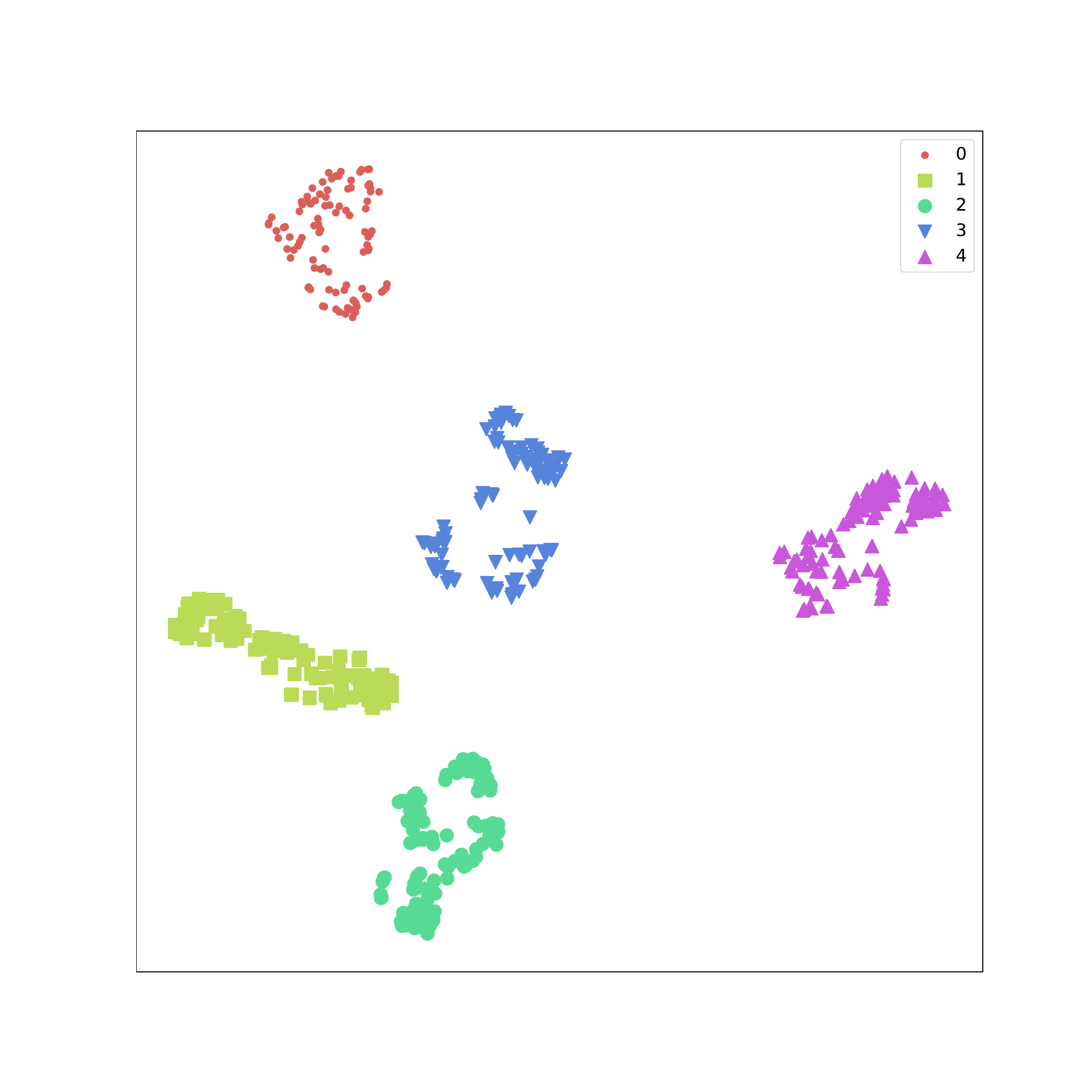}}
 \hfill 	
  \subfloat[Task 2 with \name]{\includegraphics[width=0.25\textwidth]{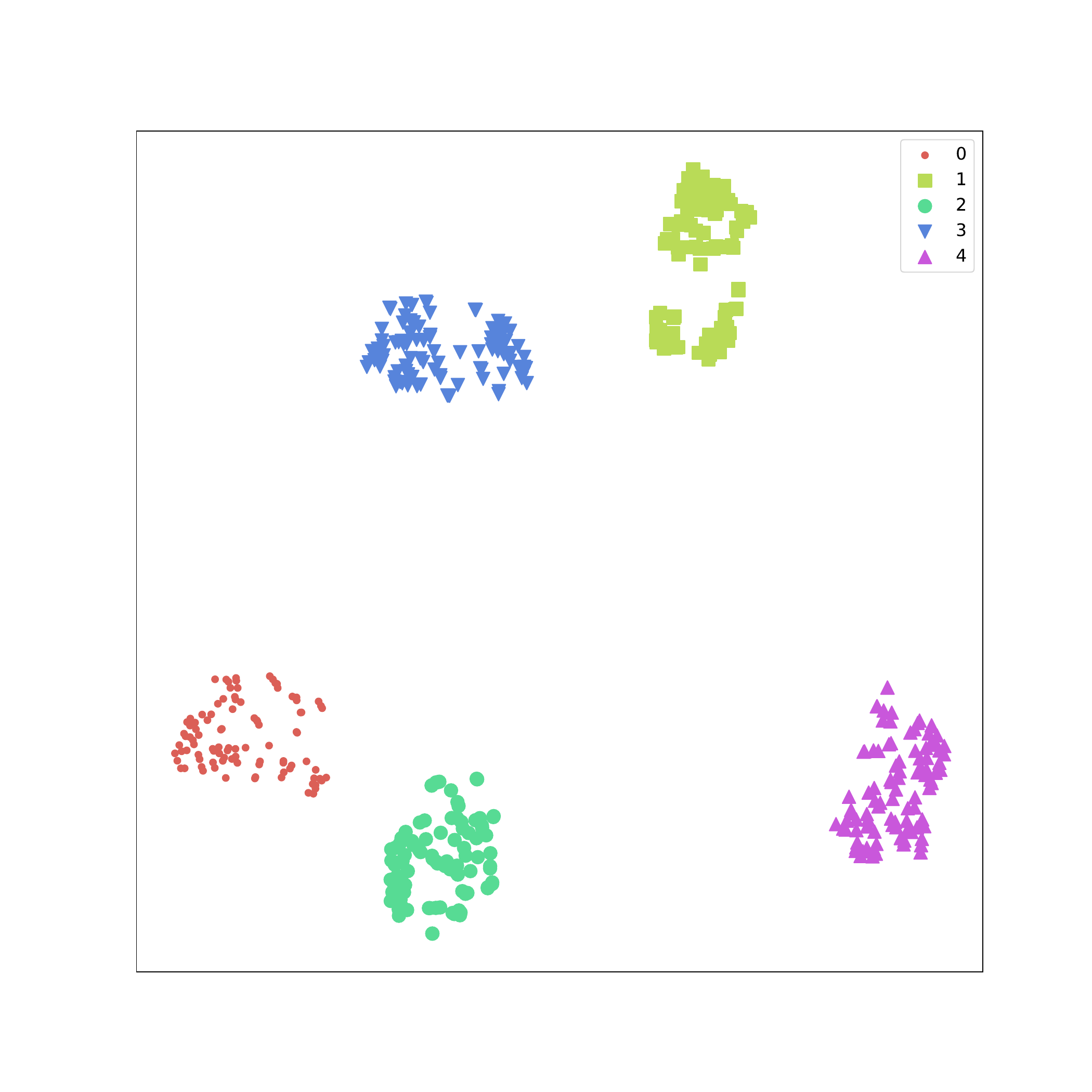}}
 \hfill	
  \subfloat[Task 3 with \name]{\includegraphics[width=0.25\textwidth]{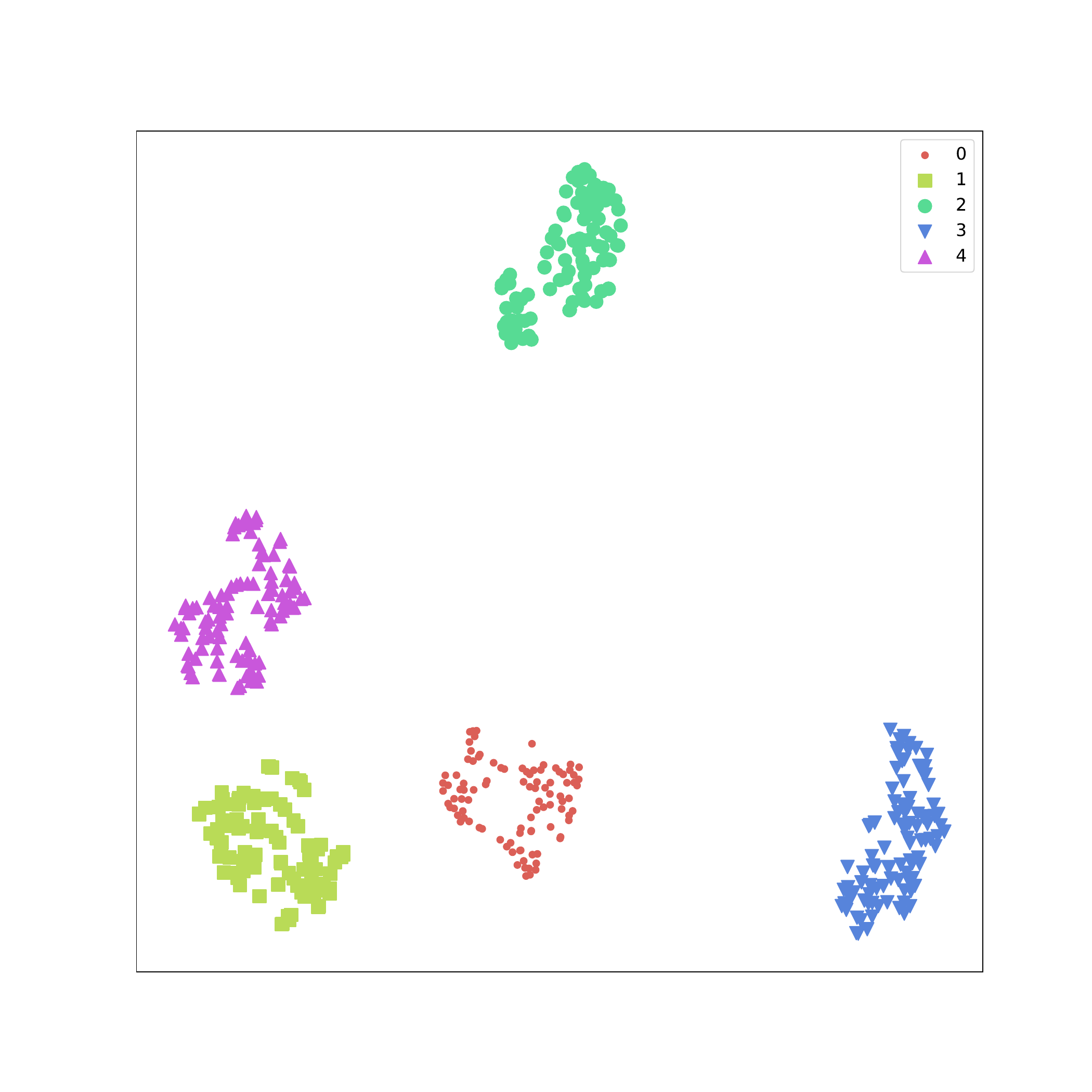}}
  \hfill	
  \subfloat[Task 4 with \name]{\includegraphics[width=0.25\textwidth]{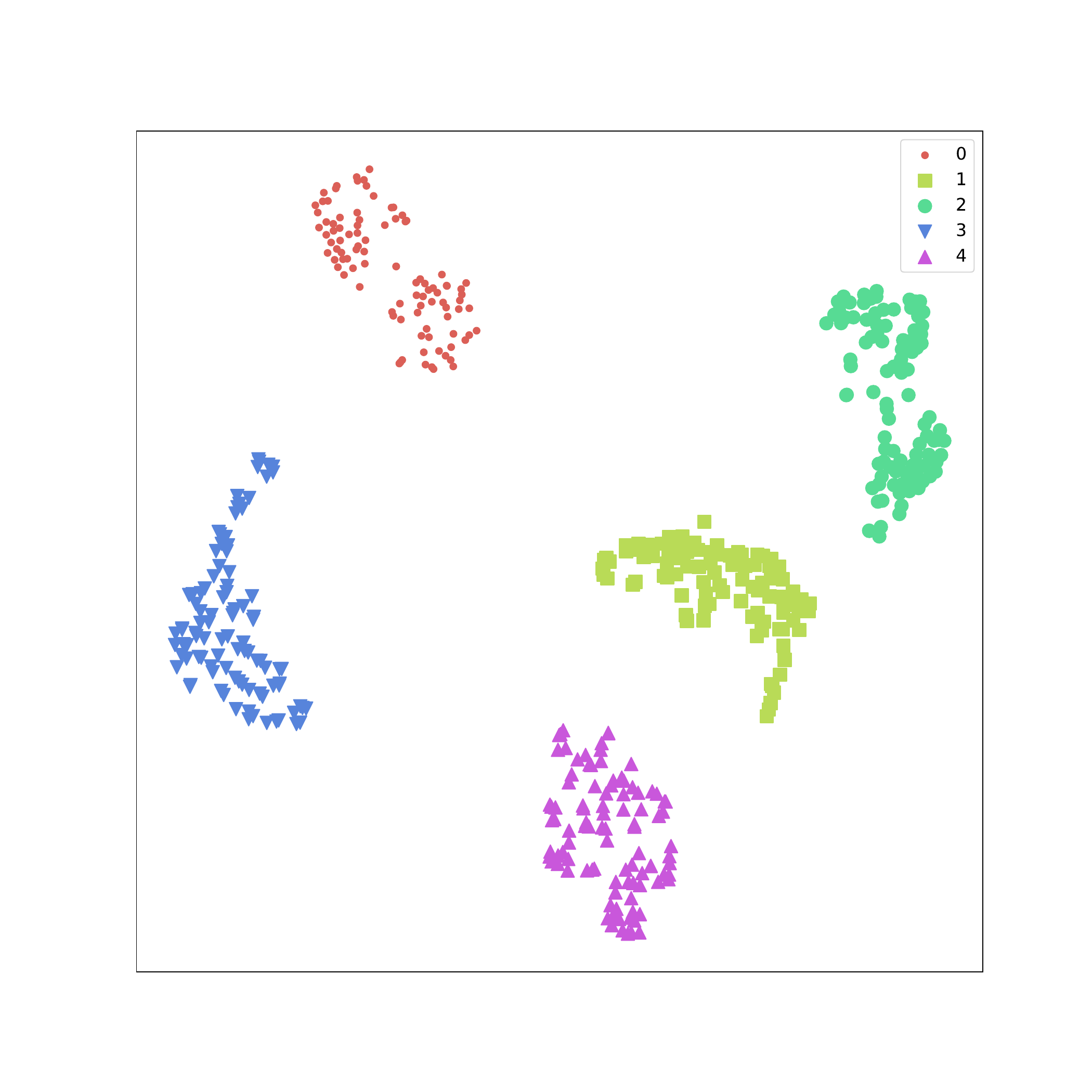}}
\caption{Visualization of class-relevant patch embedding selection for four randomly sampled 5-way 1-shot classification tasks. (a), (b), (c), and (d) show the visualizations without \name. (e), (f), (g), and (h) show the corresponding visualizations with \name. \name selects class-relevant patch embeddings by class embedding, thus eliminates class-irrelevant patch embeddings.}
\label{fig:tsne}
\end{figure*}

% Based on the patch embedding selection method mentioned above, we obtain the selected image representation sets for a query and a support image, respectively. Figure~\ref{fig:tsne} shows the patch embedding visualization results of four randomly sampled 5-way 1-shot classification tasks with/without \name. It can be observed that with \name, the selected patch embeddings in the cluster are more compact, which means it is easier to distinguish patch embeddings among different categories of images.
Using the patch embedding selection method described above, we generate the selected image representation sets for both query and support images. Figure~\ref{fig:tsne} presents the visualization results of patch embeddings for four randomly sampled 5-way 1-shot classification tasks, comparing cases with or without \name. As depicted in the figure, when \name is adopted, the selected patch embeddings within each cluster exhibit a more compact arrangement. This indicates that it becomes easier to differentiate patch embeddings among different image categories.
\begin{align}\label{e7}
    P_{selected} = [class^p,patch_{id_{1}}^p,patch_{id_{2}}^p,...,patch_{id_{m}}^p], 
\end{align}
\begin{align}\label{e8}
    Q_{selected} = [class^q,patch_{id_{1}}^q,patch_{id_{2}}^q,...,patch_{id_{m}}^q]. 
\end{align}

Once the class-relevant patch embeddings are selected, they are further fused with the class embedding to ensure that each patch feature is associated with global information while preserving the local characteristics. In this paper, we employ the fusion strategy of linear addition, which has been shown to be effective in previous works~\cite{mixer,cpea}. The resulting fused patch embeddings are as follows:
\begin{align}\label{e9}
    \overline{patch}_{id_{i}} = patch_{id_{i}} + 2class.
\end{align}

Here, we adopt a set of fusion proportions that have been validated as optimal in prior work~\cite{cpea}.
After integration, the final representation sets for both the support and query images are as follows: 
\begin{align}\label{e10}
    \overline{P_{selected}} = [\overline{patch}_{id_{1}}^p,\overline{patch}_{id_{2}}^p,...,\overline{patch}_{id_{m}}^p],
\end{align}
\begin{align}\label{e11}
    \overline{Q_{selected}} = [\overline{patch}_{id_{1}}^q,\overline{patch}_{id_{2}}^q,...,\overline{patch}_{id_{m}}^q],
\end{align}
where $m$ is the number of selected patch embeddings.

% Through the aforementioned operations, the model gains the ability to discern critical information within the focal %target
% region, thereby enhancing the representation of both support and query images while mitigating the negative impact of non-target regions. Moreover, this selection method maintains simplicity and lightweightness without introducing additional learnable parameters, ensuring consistency in selection capability across training and testing phases.

Through the aforementioned operations, the model acquires the capability to identify critical information within the focal region, thereby improving the representation of both support and query images while mitigating the negative impact of non-target regions. Furthermore, this selection method remains simple and lightweight without introducing additional learnable parameters, ensuring consistency in selection capability across both training and testing phases.

By strengthening the image representation, the margin between instances within the same class and those across different classes increases. This, in turn, leads to improved accuracy in classifying queries.

\subsection{Classification using Selected Patch Embeddings}
To calculate the similarity between the query image and the support image, we adopt a dense score matrix $S$ whose element is a score between selected patch embeddings across images, which can be formulated as follows:
\begin{align}\label{e12}
    S_{ij}(q,p) = cos(\overline{patch}_{id_{i}}^q,\overline{patch}_{id_{j}}^p)^{2}.
\end{align} 
Then, we flatten matrix $S$ and directly feed it into a multi-layer perceptron to output a similarity score.
\begin{align}\label{e13}
    score(q,p) = MLP(Flatten(S(p,q))),
\end{align}
where $MLP$ is a multi-layer perceptron. $Flatten$ reshapes metric $S$ from two dimensions to one dimension.

Therefore, a query point $\mathbf{x}$ is passed through the feature extractor to obtain the corresponding feature representation. The similarity score between $\mathbf{x}$ and the support set prototypes is calculated to obtain the probability that $\mathbf{x}$ belongs to class $n$:
\begin{align}\label{e9}
    p(y=n|\mathbf{x}) = \frac{exp(score(q_\mathbf{x},p_n ))}{\sum_{n=1}^{N}exp(score(q_\mathbf{x},p_n))}.
\end{align}
% where $cos(\cdot,\cdot)$ denotes the cosine similarity between two vectors. 
In the end, the cross entropy $\mathcal{L}$ is employed as the loss function.

\section{Experiments}
 In this section, we present the experimental results of our model, where we evaluate its characteristics and compare the performance against several baseline methods on four widely used benchmark datasets. 
We also extend our approach to existing methods and conduct an ablation study to demonstrate the effectiveness of our proposed method. Finally, we show the visualization of the image after the class-relevant patch embedding selection.

\subsection{Datasets}
In the standard few-shot classification task, there are four popular benchmark datasets for few-shot classification, including $mini$ImageNet~\cite{MiniImageNet}, $tiered$ImageNet~\cite{tieredImageNet}, CIFAR-FS~\cite{cifar-fs} and FC100~\cite{fc100} .

The $mini$ImageNet dataset contains 100 categories sampled from ILSVRC-2012~\cite{ImageNet}. Each category contains 600 images, and 60,000 images in total. In standard setting, it is randomly partitioned into training, validation and testing sets, each containing 64, 16 and 20 categories.

The $tiered$ImageNet dataset is also from ILSVRC-2012, but with a more extensive data scale. 
There are 34 super-categories, partitioned into training, validation and testing sets, with 20, 6 and 8 super-categories, respectively. It contains 608 categories, corresponding to 351, 97 and 160 categories in each partitioned dataset. 

Both CIFAR-FS and FC100 are derived from CIFAR-100, which comprises 100 classes and 600 images per class, respectively. %Unlike the previous datasets, 
CIFAR-FS and FC100 are characterized by small-resolution images, with dimensions of 32 × 32 pixels. Specifically, CIFAR-FS is randomly divided into 64 training classes, 16 validation classes, and 20 testing classes. On the other hand, FC100 contains 100 classes sourced from 36 super-classes in CIFAR100. These 36 super-classes are further divided into 12 training super-classes (containing 60 classes), 4 validation super-classes (containing 20 classes), and 4 testing super-classes (containing 20 classes).

\subsection{Implementation Details}
Motivated by the scalability and effectiveness of pre-training techniques, we employ an MIM-pretrained Vision Transformer as our model backbone. Specifically, we utilize the ViT-Small architecture with a patch size of 16.
In the pre-training phase, we employ the same strategy of \cite{FewTrue} to pretrain our ViT-Small backbones and mostly stick to the hyperparameter settings reported in their work. 
In the meta training phase, we utilize the AdamW optimizer with default settings. The initial learning rate is set to $1e-5$ and decays to $1e-6$ following a cosine learning rate scheme. %schedule. 
For the $mini$ImageNet and $tiered$ImageNet datasets, we resize the images to a size of 224 and train for 100 epochs, with each epoch consisting of 600 episodes. Similarly, for the CIFAR-FS and FC100 datasets, we resize the images to a size of 224 and train for 90 epochs, with 600 episodes each. During the fine-tuning, we adopt episode training mechanism. To be consistent with prior studies, we employ the validation set to select the best performing models.
In addition, we apply standard data augmentation techniques, including random resized loops and horizontal flips.

During the meta-testing phase, we randomly sample 1000 tasks, each containing 15 query images per class. We report the mean accuracy along with the corresponding $95\%$ confidence interval.

\subsection{Performance Comparison}

\begin{table*}[ht]
\caption{%Comparison with the state-of-the-art 
Comparisons on 5-way 1-shot and 5-way 5-shot with 95$\%$ confidence intervals on $mini$Imagenet and $tiered$Imagenet. 
} 
\label{tab:miniImage}
\begin{center}
% \scalebox{1.0}{
    \begin{tabular}{l|cr|cc|cc}
    \hline
    \multirow{2}{*}{Model} &\multirow{2}{*}{Backbone} & \multirow{2}{*}{$\approx$Params} & \multicolumn{2}{c|}{$mini$ImageNet} & \multicolumn{2}{c}{$tiered$ImageNet} \\
                           &                  &              &1-shot& 5-shot & 1-shot& 5-shot \\ 
    \hline
    \hline
    ProtoNet~\cite{PrototypicalNetwork}  & ResNet-12 & 12.4 M & 62.29±0.33  & 79.46±0.48 & 68.25±0.23  & 84.01±0.56   \\
    FEAT~\cite{FEAT} & ResNet-12 & 12.4 M &66.78±0.20 &82.05±0.14 &70.80±0.23 &84.79±0.16 \\
    CAN~\cite{CAN} & ResNet-12 & 12.4 M& 63.85±0.48  & 79.44±0.34 &69.89±0.51 &84.23±0.37 \\
    CTM~\cite{CTM} & ResNet-18 & 11.2 M& 64.12±0.82 & 80.51±0.13 &68.41±0.39 &84.28±1.73 \\
    ReNet~\cite{RENet} & ResNet-12 & 12.4 M& 67.60±0.44  & 82.58 ±0.30 &71.61±0.51 &85.28±0.35  \\
    DeepEMD~\cite{deepemd} & ResNet-12 &12.4 M &65.91±0.82 &82.41±0.56 &71.16±0.87 &86.03±0.58 \\
    IEPT~\cite{IEPT} &ResNet-12 &12.4 M & 67.05±0.44 & 82.90±0.30 & 72.24±0.50 & 86.73±0.34 \\
    MELR~\cite{MELR} &ResNet-12 &12.4 M & 67.40±0.43 & 83.40±0.28 & 72.14±0.51 & 87.01±0.35 \\
    FRN~\cite{FRN} &ResNet-12 &12.4 M & 66.45±0.19 & 82.83±0.13 & 72.06±0.22 & 86.89±0.14 \\
    CG~\cite{CG/CNL} &ResNet-12 &12.4 M & 67.02±0.20 & 82.32±0.14 & 71.66±0.23 & 85.50±0.15 \\
    DMF~\cite{DMF} &ResNet-12 &12.4 M & 67.76±0.46 & 82.71±0.31 & 71.89±0.52 & 85.96±0.35 \\
    InfoPatch~\cite{InfoPatch} &ResNet-12 &12.4 M & 67.67±0.45 & 82.44±0.31 & - & -\\
    BML~\cite{BML} &ResNet-12 &12.4 M & 67.04±0.63 & 83.63±0.29 & 68.99±0.50 & 85.49±0.34 \\
    CNL~\cite{CG/CNL} &ResNet-12 &12.4 M & 67.96±0.98 & 83.36±0.51 & 73.42±0.95 & 87.72±0.75 \\
    Meta-NVG~\cite{Mata-NVG} &ResNet-12 & 12.4 M & 67.14±0.80 & 83.82±0.51 & 74.58±0.88 & 86.73±0.61 \\ 
    PAL~\cite{PAL} &ResNet-12 & 12.4 M & 69.37±0.64 & 84.40±0.44 & 72.25±0.72 & 86.95±0.47 \\
    COSOC~\cite{COSOC} &ResNet-12 & 12.4 M & 69.28±0.49 & 85.16±0.42 & 73.57±0.43 & 87.57±0.10 \\
    Meta DeepBDC~\cite{Meta-DeepBDC} &ResNet-12 & 12.4 M & 67.34±0.43 & 84.46±0.28 & 72.34±0.49 & 87.31±0.32 \\
    QSFormer~\cite{qsformer} &ResNet-12 & 12.4 M & 65.24±0.28 & 79.96±0.20 & 72.47±0.31 & 85.43±0.22 \\
    \hline
    LEO~\cite{LEO} & WRN-28-10 & 36.5 M & 61.76±0.08 & 77.59±0.12 & 66.33±0.05 & 81.44±0.09 \\
    CC+rot~\cite{CC-rot} & WRN-28-10 & 36.5 M & 62.93±0.45 & 79.87±0.33 & 70.53±0.51 & 84.98±0.36 \\
    FEAT~\cite{FEAT} & WRN-28-10 & 36.5 M & 65.10±0.20 & 81.11±0.14 & 70.41±0.23 & 84.38±0.16 \\
    % PSST~\cite{PSST} & WRN-28-10 & 36.5 M & 64.16±0.44 & 80.64±0.32 & - & - \\
    MetaQDA~\cite{MetaQDA} & WRN-28-10 & 36.5 M & 67.83±0.64 & 84.28±0.69 & 74.33±0.65 & 89.56±0.79 \\ 
    OM~\cite{OM} & WRN-28-10 & 36.5 M & 66.78±0.30 & 85.29±0.41 & 71.54±0.29 & 87.79±0.46 \\
    \hline
    FewTURE~\cite{FewTrue}  & ViT-Small & 22 M & 68.02±0.88 &  84.51±0.53 & 72.96±0.92 &  86.43±0.67\\
    FewTURE~\cite{FewTrue}  & Swin-Tiny & 29 M & 72.40±0.78 &  86.38±0.49 & 76.32±0.87 & 89.96±0.55\\
    CPEA~\cite{cpea} & ViT-Small & 22 M & 71.97±0.65 & 87.06±0.38 & 76.93±0.70 & 90.12±0.45\\
    \hline
    % IMAformer (ours) & ViT-Small & 22 M & \emph{70.25±0.61}  &  \emph{86.48±0.44}  & \emph{75.15±0.67}  &  \emph{88.44±0.65} \\    % two layers       
    % IMAformer (ours) & ViT-Small & 22 M & \emph{78.48±0.39}  &  \emph{89.05±0.28}  & \emph{77.70±0.67}  &  \emph{90.68±0.65} \\
     \name (ours) & ViT-Small & 22 M & \bf73.62±0.59  & \bf88.61±0.35  & \bf78.25±0.70  & \bf91.25±0.41 \\
    \hline
    \end{tabular}
\end{center}

\end{table*}

\begin{table*}[ht!]
\caption{%Comparison with the state-of-the-art 
Comparisons on 5-way 1-shot and 5-way 5-shot  with 95$\%$ confidence intervals on CIFAR-FS and FC100.}
\label{tab:cifar}
\begin{center}
% \scalebox{1.0}{
    \begin{tabular}{l|cr|cc|cc}
    \hline
    % \cline{1-1} \cline{3-5}
    \multirow{2}{*}{Model} &\multirow{2}{*}{Backbone} & \multirow{2}{*}{$\approx$Params} & \multicolumn{2}{c|}{CIFAR-FS} & \multicolumn{2}{c}{FC100} \\
             &          &            & 1-shot           & 5-shot         & 1-shot & 5-shot      \\ 
    \hline
    \hline
    ProtoNet~\cite{PrototypicalNetwork} & ResNet-12 & 12.4 M & - & -  &41.54±0.76 & 57.08±0.76 \\
    MetaOpt~\cite{MetaOpt} & ResNet-12 & 12.4 M & 72.00±0.70 & 84.20±0.50 & 41.10±0.60 & 55.50±0.60 \\
    MABAS~\cite{MABAS} & ResNet-12 & 12.4 M & 73.51±0.92 & 85.65±0.65 & 42.31±0.75 & 58.16±0.78 \\
    RFS~\cite{RFS} & ResNet-12 & 12.4 M & 73.90±0.80 & 86.90±0.50 & 44.60±0.70 & 60.90±0.60 \\
    BML~\cite{BML} & ResNet-12 & 12.4 M & 73.45±0.47 & 88.04±0.33 & -& - \\
    CG~\cite{CG/CNL} & ResNet-12 & 12.4 M & 73.00±0.70 &85.80±0.50 & -& - \\
    Meta-NVG~\cite{Mata-NVG} & ResNet-12 & 12.4 M & 74.63±0.91 & 86.45±0.59 & 46.40±0.81 & 61.33±0.71 \\ 
    RENet~\cite{RENet} & ResNet-12 & 12.4 M & 74.51±0.46 & 86.60±0.32 & - & - \\
    TPMN~\cite{TPMN} & ResNet-12 & 12.4 M & 75.50±0.90 & 87.20±0.60 & 46.93±0.71 & 63.26±0.74 \\
    MixFSL~\cite{MixFSL} & ResNet-12 & 12.4 M & -& -& 44.89±0.63 & 60.70±0.60 \\
    QSFormer~\cite{qsformer} &ResNet-12 & 12.4 M & - & - & 46.51±0.26 & 61.58±0.25 \\
    \hline
    CC+rot~\cite{CC-rot} & WRN-28-10 & 36.5 M & 73.62±0.31 & 86.05±0.22 & -& - \\
    PSST~\cite{PSST} & WRN-28-10 & 36.5 M & 77.02±0.38 & 88.45±0.35 & -& - \\
    Meta-QDA~\cite{MetaQDA} & WRN-28-10 & 36.5 M & 75.83±0.88 & 88.79±0.75 & - & -\\
    \hline
    FewTURE~\cite{FewTrue}  & ViT-Small & 22 M & 76.10±0.88 & 86.14±0.64 &46.20±0.79 &63.14±0.73 \\
    FewTURE~\cite{FewTrue}  & Swin-Tiny & 29 M & 77.76±0.81 &88.90±0.59 &47.68±0.78 &63.81±0.75 \\
    CPEA~\cite{cpea} & ViT-Small & 22 M & 77.82±0.66 & 88.98±0.45 & 47.24±0.58 & 65.02±0.60 \\
    \hline
    \name (ours) & ViT-Small & 22 M & \bf78.96±0.63	& \bf90.63±0.42	& \bf48.72±0.63	& \bf66.55±0.59 \\
    \hline
    \end{tabular}
\end{center}

\end{table*}

Following the established conventions of few-shot learning, we perform experiments on four popular few-shot classification benchmarks, and the results are presented in Tables~\ref{tab:miniImage} and ~\ref{tab:cifar}, respectively. 
From the results, it is evident that our proposed \name consistently achieves competitive performance compared to the state-of-the-art methods on both the 5-way 1-shot and 5-way 5-shot tasks.

\textbf{Results on $mini$Imagenet and $tiered$Imagenet datasets}. Table~\ref{tab:miniImage} presents a comparison on the 1-shot and 5-shot performance of our method with the baselines on the $mini$Imagenet and $tiered$Imagenet datasets. We achieve significant improvements over existing  SOTA results while utilizing fewer learnable parameters. For instance, on the $mini$Imagenet dataset, our \name surpasses the runner-up by 1.22\% in the 1-shot setting and 1.66\% in the 5-shot setting. Similarly, on the $tiered$Imagenet dataset, \name outperforms the runner-up  by 1.32\% and 1.13\% in the 1-shot and 5-shot settings, respectively.

Compared to FewTURE and CPEA, which utilize all patch embeddings, our method eliminates the class-irrelevant patch embeddings while achieving better performance and utilizing fewer learnable parameters. The significant margin between our method and the runner-up baselines further validates the contribution of our method, which effectively captures key information in the data and maintains good generalization capability.

\textbf{Results on small-resolution datasets}. To verify the adaptability of the model, we conduct further experiments on two small-resolution datasets along with comparisons to other methods. This allows us to evaluate the performance of our method across various data scenarios, ensuring fair and comprehensive performance analysis.

Table~\ref{tab:cifar} demonstrates the 1-shot and 5-shot classification performance on the small-resolution datasets CIFAR-FS and FC100. In the case of CIFAR-FS, \name outperforms the 
runner-up by 1.14\% in the 1-shot setting and 1.65\% in the 5-shot setting. Similarly, for FC100, our method surpasses the runner-up  by 1.04\% in the 1-shot setting and 1.53\% in the 5-shot setting.

These results highlight the effectiveness of our method across all the four datasets, since our method achieves %state-of-the-art 
superior performance across all settings.

The substantial enhancements observed affirm the superiority of our method, which employs the class-relevant patch embedding selection method to augment the representation of an image. The effectiveness of this approach is substantiated by the remarkable performance gains achieved across diverse few-shot learning tasks.

\subsection{Model Extension}
Our \name method is a flexible strategy. We can consider inserting our method as long as all patch embeddings are used to measure the similarity between images. Here, we extend our approach to the following two methods: FewTURE~\cite{FewTrue} and CPEA~\cite{cpea}. FewTURE first extracts the class embeddings and patch embeddings of images using pre-trained ViT and then learns a patch weight by using all patch embeddings to determine which patches are helpful for classification. For this method, we first use class embedding to select the patch embedding related to the category and then conduct subsequent operations based on the selected patches. 
CPEA also uses pre-trained ViT to extract image class embedding and patch embedding and then fuses all patch embeddings with class embedding to perform image classification measurement.
Similarly, we select the patches related to the class and proceed to the following operations. The experimental results are shown in Table~\ref{tab:extension}. After adding our patch selection method, the effects of these two methods are further improved by approximately 1.5\%, which once again proves the effectiveness and flexibility of our method.

\begin{table}[h!]
\caption{%Impact of 
Results when integrating our method into existing methods for few-shot classification on $mini$ImageNet. 
% We show the average 5-way accuracy with 95$\%$ confidence interval.
}
\label{tab:extension}
\begin{center}
% \scalebox{1.0}{
    \begin{tabular}{l|c|c}
    \hline
    % Number of selected & \multirow{2}{*}{1-shot} & \multirow{2}{*}{5-shot} \\
    Model    &    1-shot       &    5-shot     \\ 
    \hline
    \hline
   FewTURE~\cite{FewTrue} & 68.02±0.88  & 84.51±0.53 \\
   FewTURE+\name & 69.66±0.84  & 85.94±0.52 \\
   CPEA~\cite{cpea} & 71.97±0.65  & 87.06±0.38 \\
   CPEA+\name & 73.79±0.66  & 88.52±0.35 \\ 
    \hline
    \end{tabular}
\end{center}

\end{table}

\subsection{Ablation Study}
In this subsection, we conduct ablation experiments to analyze the impact of each component on the performance of our method. Specifically, we focus on the 5-way 5-shot setting on the $mini$Imagenet dataset, utilizing the pre-trained Vision Transformer backbone. By analyzing the performance changes resulting from these ablations, we gain insights into the contribution of each component to the overall effectiveness of our method. Specifically, we investigate the influence of the following variations.

\textbf{Class-relevant Patch Embedding Selection method}: We evaluate the performance of our model without class-relevant patch embedding selection, which is responsible for enhancing the representation of the support and query sets.

To verify the effectiveness of our class-relevant patch embedding selection method, we design a comparative model called Vanilla. In the Vanilla method, we do not employ patch selection, and all the patch embeddings of the support and query sets are utilized to measure the similarity score. 
As shown in Table~\ref{tab:method}, the results clearly demonstrate the significant improvement achieved by our \name method compared to the Vanilla method. 
This result confirms the efficacy of our selection method in improving the performance for few-shot learning tasks.

\begin{table}[h!]
\caption{Results of the class-relevant patch embedding selection method for few-shot classification on $mini$ImageNet. 
% We show the average 5-way accuracy with 95$\%$ confidence interval.
}
\label{tab:method}
\begin{center}
% \scalebox{1.0}{
    \begin{tabular}{c|c|c}
    \hline
    Patch embedding selection    &     1-shot      &      5-shot   \\ 
    \hline
    \hline
   $\surd$  & 73.62±0.59  & 88.61±0.35 \\
   $\times$ & 70.74±0.69  & 86.69±0.38 \\
    \hline
    \end{tabular}
\end{center}
\end{table}

\textbf{Number of selected patch embeddings}: We evaluate the performance of our model with different numbers of selected patch embeddings, which is the dominant factor in the selection method.

We design a series of comparative models to verify the effectiveness of a different number of selected patch embeddings. $m=0$ means that the class embeddings are directly treated as image representations, while $m=196$ means that all the patch embeddings are treated as image representations.
As shown in Table~\ref{tab:number}, the performance degrades when the selected patch embeddings are either too small or too large. We speculate this is because when too few patch embeddings are selected, the target region cannot be effectively included. However, when too many patch embeddings are selected, the image representation will be affected because too many class-irrelevant patches are retained. In this paper, we set the number of selected patch embeddings to 96 as a good tradeoff. 

\begin{table}[h!]
\caption{Results on the number of selected class-relevant patch embeddings for few-shot classification on $mini$ImageNet. 
% We show the average 5-way accuracy with 95$\%$ confidence interval.
}
\label{tab:number}
\begin{center}
% \scalebox{1.0}{
    \begin{tabular}{c|c|c}
    \hline
    Number of selected & \multirow{2}{*}{1-shot} & \multirow{2}{*}{5-shot} \\
    patch embeddings ($m$)    &           &         \\ 
    \hline
    \hline
   0 & 70.47±0.66  & 85.18±0.55 \\
   49 & 72.17±0.66  & 87.34±0.35 \\
   64 & 73.06±0.64  & 88.44±0.35 \\
   96 & 73.62±0.59  & 88.61±0.35 \\ 
   128 & 72.80±0.65  & 87.80±0.35 \\
   196 & 70.74±0.69  & 86.69±0.38 \\
    \hline
    \end{tabular}
\end{center}
\end{table}

\begin{table}[h!]
\caption{Analysis on different distance function employed in class-relevant patch embedding selection for few-shot classification on $mini$ImageNet. 
% We show the average 5-way accuracy with 95$\%$ confidence interval.
}
\label{tab:distance}
\begin{center}
% \scalebox{1.0}{
    \begin{tabular}{c|c|c}
    \hline
    Different distance function & \multirow{2}{*}{1-shot} & \multirow{2}{*}{5-shot} \\
    in \name    &           &         \\ 
    \hline
    \hline
   $dot(\cdot)$ & 72.52±0.65  & 88.12±0.37 \\
   $abs(\cdot)$ & 72.53±0.65  & 88.19±0.37 \\
   $sqr(\cdot)$ & 72.55±0.65  & 88.11±0.37 \\ 
   $cos(\cdot)$ & 73.62±0.59  & 88.61±0.35 \\ 
    \hline
    \end{tabular}
\end{center}
\end{table}

\begin{table}[h!]
\caption{Analysis on different location of class-relevant patch embedding selection for few-shot classification on $mini$ImageNet. 
% We show the average 5-way accuracy with 95$\%$ confidence interval.
}
\label{tab:location}
\begin{center}
% \scalebox{1.0}{
    \begin{tabular}{c|c|c}
    \hline
    Location of & \multirow{2}{*}{1-shot} & \multirow{2}{*}{5-shot} \\
    \name    &           &         \\ 
    \hline
    \hline
   12 & 73.62±0.59  & 88.61±0.35 \\
   11 & 72.30±0.65  & 87.90±0.37 \\
   10 & 72.14±0.65  & 87.67±0.37 \\
   9 & 72.01±0.65  & 87.38±0.38 \\ 
   8 & 71.83±0.66  & 86.98±0.39 \\
   7 & 71.51±0.66  & 86.96±0.38 \\
   6 & 71.25±0.65  & 86.56±0.39 \\
    \hline
    \end{tabular}
\end{center}
\end{table}

\textbf{Distance function employed in \name}: 
In order to determine the most suitable metric for selecting class-relevant patch embeddings, we examined four metric methods: dot product ($dot$), Manhattan distance ($abs$), Euclidean distance ($sqr$), and cosine similarity ($cos$).
Table~\ref{tab:distance} presents the impact of these different distance functions in selecting class-relevant patch embeddings. The experiments demonstrate that cosine similarity, which is chosen in our method, yields the most prominent results.
%why we chose it for this paper.

\textbf{Location of \name}:
The Vision Transformer typically comprises an encoder with twelve layers, and the placement of our \name module at different locations within this architecture can yield varying effects.
Table~\ref{tab:location} illustrates the impact of different locations for selecting class-relevant patch embeddings on few-shot classification performance. The model achieved its best results when the \name module was positioned after the final layer, with performance declining as the module was placed earlier in the network. We speculate that this trend arises because the semantic information of class and patch embeddings obtained after the final layer is more accurate, making it more conducive in selecting class-relevant patch embeddings. In this paper, we position the module after the final layer of the Vision Transformer.

\begin{figure*}[t!]
\begin{center}
  \includegraphics[width=1.0\linewidth]{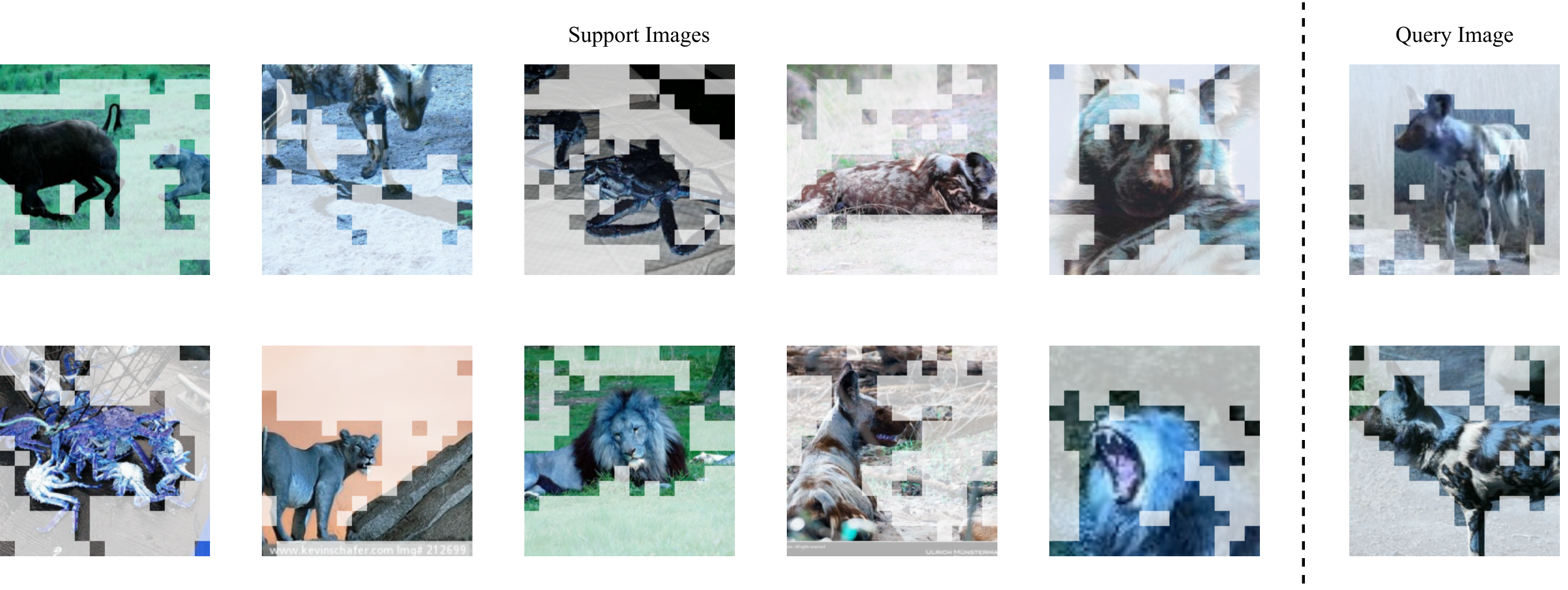}
\end{center}
   \caption{
   Illustration of the selected patch embeddings visualization of two randomly sampled 5-way 1-shot classification tasks with one query image per class. The selected class-relevant patches are retained while the class-irrelevant patches are masked. One can observe that the selected patches mainly focus on the focal region.}
\label{fig:mask}
\end{figure*}

\subsection{Visualization Analysis}
To qualitatively assess the efficacy of the proposed method for selecting class-relevant patch embeddings, we present visualization results of query and support set image patches using our model. As depicted in Figure ~\ref{fig:mask}, the selected patches highlight the target region of the image, retaining relevant features while eliminating interference from regions such as the background.

This visualization underscores the effectiveness of our approach. By selecting class-relevant patch embeddings prior to measurement, the resulting query and support features exhibit greater distinguishability.

\section{Conclusion}
In summary, our paper proposed a novel method called \name, 
which utilizes the class embedding to select class-relevant patch embeddings in the context of few-shot learning. 
%Our approach involves segmenting the input image into smaller patches and encoding these local patches using a pre-trained Vision Transformer architecture, achieved through self-supervised learning. This segmentation allows us to obtain both global information representation through the class embedding and local information representation through the patch embeddings. 
By segmenting input images into smaller patches and encoding them using a pre-trained Vision Transformer architecture learned through self-supervised learning, our approach captures both global and local information representations. 
%We introduced a selecting class-relevant patch embeddings method to enhance the representations of support and query sets. This method enables the support set images and the query image to focus on the target region presented in the image, thereby enhancing its representations and promoting similarity between instances. 
The introduction of a method for selecting class-relevant patch embeddings enhances the representations of support and query sets, enabling them to focus on focal point and promoting similarity between instances.

Empirical evaluations demonstrated the effectiveness of our proposed method, yielding advanced performance. 
Our approach offers several advantages:
%Firstly, it significantly improves the image representation by selecting class-relevant patch embeddings. 
%Secondly, it efficiently utilizes the knowledge and representations encoded in these pre-trained models learned from Masked Image Modeling pre-training task. 
it significantly enhances image representation by selecting class-relevant patch embeddings and efficiently utilizes knowledge and representations encoded in pre-trained models from the Masked Image Modeling pre-training task. 
By building upon these pre-trained models, our method has the potential to achieve competitive performance in few-shot learning tasks without the need of  external weight modules. 
% ---------------------------------------------------------------------------------

%\section*{Acknowledgments}
%This work is supported by National Natural Science Foundation (U22B2017).
% ---------------------------ref----------------
\bibliographystyle{IEEEtran}
\bibliography{References}

\newpage

 

\begin{IEEEbiography}[{\includegraphics[width=1in,height=1.25in,clip,keepaspectratio]{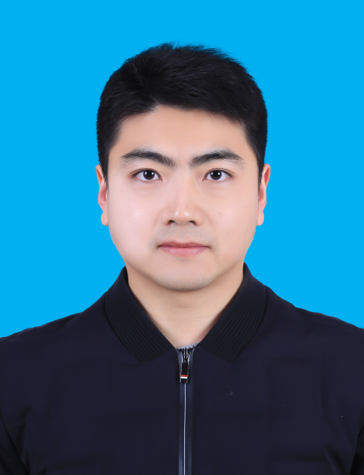}}]{Weihao Jiang}
is currently pursuing a Ph.D. degree in School of Computer Science and Technology, Huazhong University of Science and Technology, Wuhan, China.
His research focuses on deep learning,
few-shot learning and meta-learning.
\end{IEEEbiography}

\vspace{11pt}

\begin{IEEEbiography}[{\includegraphics[width=1in,height=1.25in,clip]{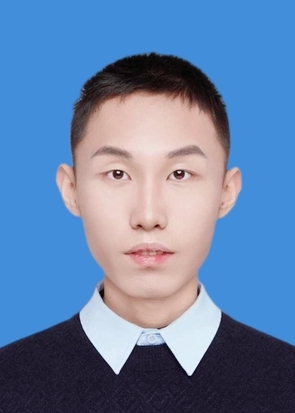}}]{Haoyang Cui}
is currently pursuing a undergraduate degree in School of Computer Science and Technology, Huazhong University of Science and Technology, Wuhan, China. 
His research mainly focuses on deep learning and image classification.
\end{IEEEbiography}

\vspace{11pt}

\begin{IEEEbiography}[{\includegraphics[width=1in,height=1.25in,clip,keepaspectratio]{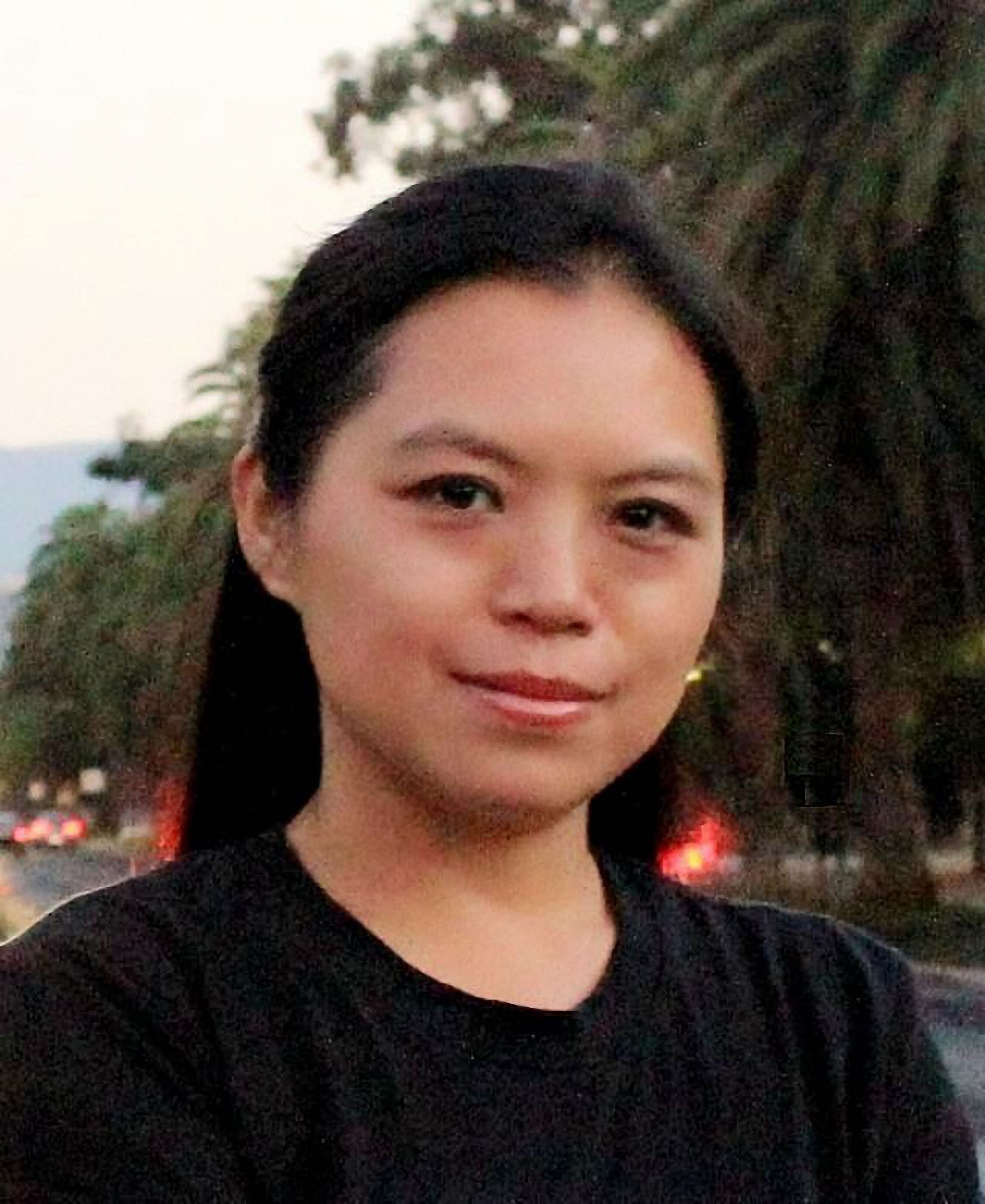}}]{Kun He}
(SM18) is currently a Professor in School of Computer Science and Technology, Huazhong University of Science and Technology, Wuhan, P.R. China. She received the Ph.D. degree in system engineering from Huazhong University of Science and Technology, Wuhan, China, in 2006. She had been with the Department of Management Science and Engineering at Stanford University in 2011-2012 as a visiting researcher. She had been with the de- partment of Computer Science at Cornell University in 2013-2015 as a visiting associate professor, in 
2016 as a visiting professor, and in 2018 as a visiting professor. She was honored as a Mary Shepard B. Upson visiting professor for the 2016-2017 Academic year in Engineering, Cornell University, New York. Her research interests include adversarial machine learning, deep representation learning, %social network analysis, and combinatorial optimization. 
graph machine learning and optimization method. 
\end{IEEEbiography}

\vfill

\end{document}